\documentclass[preprint]{elsarticle}
\usepackage{amsmath,amsfonts}
\usepackage{array}
\usepackage{textcomp}
\usepackage{stfloats}
\usepackage{url}
\usepackage{verbatim}
\usepackage{graphicx}
\usepackage{multirow}
\usepackage{bbding}
\usepackage{gensymb}
\usepackage{booktabs}
\usepackage{subfigure}
\usepackage[ruled,linesnumbered]{algorithm2e}
\usepackage{xcolor}
\usepackage{threeparttable}
\usepackage{tabularx}
\usepackage{bbm}
\usepackage{pifont}
\usepackage{color}
\usepackage{soul}
\usepackage{float}
\usepackage{lineno,hyperref}

\begin{document}

\begin{frontmatter}

\title{Learning Representations for Clustering via Partial Information Discrimination and Cross-Level Interaction}

\author[1]{Hai-Xin Zhang}
\ead{reganzhx@stu.scau.edu.cn}
\author[1]{Dong Huang}
\ead{huangdonghere@gmail.com}
\author[2]{Hua-Bao Ling}
\ead{hbling@stu.scau.edu.cn}
\author[1]{Guang-Yu Zhang}
\ead{guangyuzhg@foxmail.com}
\author[3]{Weijun~Sun}
\ead{changdongwang@hotmail.com}
\author[1]{Zihao Wen}
\ead{zihao.wen@hotmail.com}
\address[1]{College of Mathematics and Informatics, South China Agricultural University, China}
\address[2]{School of Computer Science and Engineering, Sun Yat-sen University, China}
\address[3]{School of Automation, Guangdong University of Technology, China}

\begin{abstract}
Despite the significant advances in the deep clustering research, there remain three critical limitations to most of the existing approaches. First, they mostly seek to learn the clustering result by associating some distribution-based loss to specific network layers, neglecting the potential benefits of harnessing the contrastive sample-wise relationships. Second, they usually focus on the representation learning at the full-image scale, overlooking the discriminative information latent in partial image regions. Third, while certain prior approaches can jointly perform multiple levels of learning, they frequently fail to exploit the interaction between different learning levels. To overcome these limitations, this paper presents a novel deep image clustering approach via Partial Information discrimination and Cross-level Interaction (PICI). Specifically, we utilize a Transformer encoder as the backbone, which are associated with two types of augmentations to formulate two paralleled views. The augmented samples, integrated with masked patches, are processed through the Transformer encoder to derive class tokens. Subsequently, three partial information learning module are jointly enforced, namely, the partial information self-discrimination (PISD) module which trains the network by masked image reconstruction, the partial information contrastive discrimination (PICD) module which performs the contrastive clustering at the instance-level and the cluster-level simultaneously, and the cross-level interaction (CLI) module which aims to ensure the consistency across different learning levels. Through this joint formulation, our PICI approach for the first time, to our knowledge, bridges the gap between the masked image modeling and the deep contrastive clustering, offering a novel pathway for enhanced representation learning and clustering. Extensive experimental results across six image datasets validate the superiority of our PICI approach over the state-of-the-art. The source code is available at \url{https://github.com/Regan-Zhang/PICI}.
\end{abstract}


\end{frontmatter}


\section{Introduction}
\label{sec:introduction}
Data clustering is a fundamental technique in pattern recognition and machine learning. Traditional clustering methods usually assume that some hand-crafted features are given and then partition the dataset based on these given features, which, however, lack the important ability of feature representation learning and may yield suboptimal clustering results especially for some high-dimensional complex data such as images videos. Due to its joint ability of representation and clustering, the deep learning-based clustering methods, also known as the deep clustering methods, have captured enormous attention in recent years.

Previous deep clustering methods \cite{xie2016unsupervised,guo2017improved,guo2018deep} typically employ a deep neural network to learn feature representations and further obtain the clustering result by enforcing some cluster distribution-based clustering loss. Despite the considerable progress that has been achieved, previous deep clustering methods  \cite{xie2016unsupervised,guo2017improved,guo2018deep} mostly suffer from three critical limitations. 
First, they often rely on the some global distribution-based loss (e.g., via the Kullback–Leibler (KL) divergence-based clustering loss) to learn the clustering result, but overlook the rich and contrastive information among the sample-wise relationships. Second, they mostly perform the feature learning at the full-image scale, but neglect the opportunities in discovering more discriminative semantics from only partial (or masked) regions. Third, as the instance-level modeling and the cluster-level modeling are two key factors in the deep clustering formulation, surprisingly, few previous works are able to enable the cross-level interaction to adaptively and mutually enhance the cross-level learning.

Recently some efforts have been carried out to partially address one or two of the above three limitations. To utilize the contrastive information among the sample-wise relationship, some contrastive learning-based deep clustering methods have been developed, such as Contrastive Clustering (CC) \cite{li2021contrastive}, Instance Discrimination and Feature Decorrelation (IDFD) \cite{tao2021clustering}, and Prototypical Contrastive Learning (PCL) \cite{li2020prototypical}, which aim to explore the instance-wise (i.e., sample-wise) relationships via the contrastive learning paradigm. But on the one hand,  these contrastive deep clustering methods still rely on the representation learning at the full-image scale while lacking the partial information discrimination ability. On the other hand, they tend to perform the instance-level contrastive learning and the cluster-level contrastive learning via two projectors, respectively, without considering taking advantage of the interaction between them. To train the network with partial information discrimination,  the Masked Image Modeling (MIM) \cite{he2022masked} arises as a new trend for self-supervised learning. As a representative model in MIM, the Masked Auto-Encoder (MAE) utilizes a specific task that reconstructs the masked images to improve the ability of representation learning via Vision Transformer (ViT) \cite{DosovitskiyB0WZ21}, which shows that the reconstruction and discrimination of partial image information may significant benefit the representation learning. The MAE offers a promising pathway for harnessing the partial information in images, which, however, is devised for the representation learning task only and still lacks the deep clustering ability and the cross-level contrastive learning ability.
Despite these impressive efforts, it remains an open problem how to tackle the above three limitations simultaneously and furthermore formulate the contrastive relationships, the cross-level intraction, and the partial information discrimination in a unified deep representation learning and clustering framework.

\begin{figure*}[!t]
	\centering
	\includegraphics[width=1\textwidth]{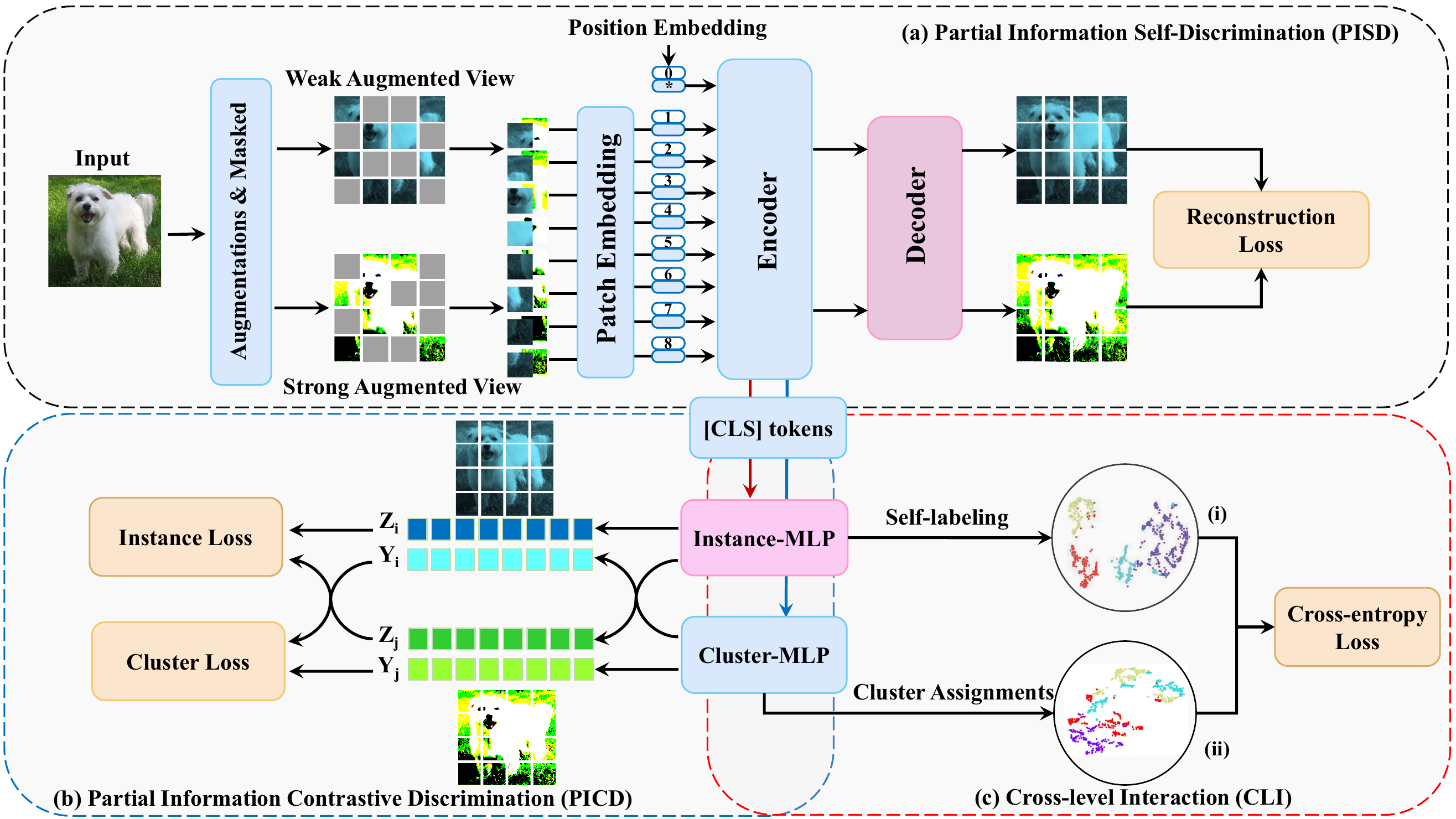}
	\caption{An overview of the proposed PICI framework, which jointly incorporates three learning modules, namely, (a) the PISD module, which enforces the partial information self-discrimination upon the masked images via the Transformer auto-encoder, (b) the PICD module, which takes the class tokens [CLS] as input and achieves the partial information contrastive information discrimination via two levels of contrastive learning, and (c) the CLI module, which enables the mutual interaction between the instance-level and cluster-level subspaces by constraining their cross-level consistency.}
	\label{fig}
\end{figure*}

To jointly address the aforementioned three limitations, in this paper, we propose a novel deep clustering framework based on Partial Information discrimination and Cross-level Interaction (PICI). Different from previous deep clustering methods that mostly utilize the convolutional neural network (CNN), we take advantage of the Transformer encoder as the backbone to capture the global relevance information via the self-attention mechanism (as shown in Fig.~\ref{fig}). Particularly, two types of augmentations are first performed on an input image to generate two augmented samples. Each of the augmented sample is split into a sequence of patches, where the patches are then randomly masked so as to incur partial information loss that plays an important role in learning discriminant and semantic information from images.  Thereafter, the two parallel views of augmented samples with masked patches are fed into the encoder (with position embedding) to produce the class tokens, denoted as [CLS]. 
Furthermore, three types of learning are jointly enforced, which respectively correspond to two modules of partial information discrimination, i.e., the partial information self-discrimination (PISD) module and the partial information contrastive discrimination (PICD) module, and a module of cross-level learning, i.e., the cross-level interaction (CLI) module. Specifically, in PISD, a decoder is utilized to reconstruct the masked patches of the augmented sample, which seeks to learn the semantic information of the image by recovering the missing local regions. In PICD, the class tokens [CLS] of each augmented view are fed to two multi-layer perceptron (MLP)-based projectors, namely, the instance-MLP and the cluster-MLP, through which the instance-level contrastive learning and the cluster-level contrastive learning can be enforced, respectively. Notice that the two levels of contrastive learning are performed separately. To bridge the gap between the two levels of contrastive learning, the cross-level adaptive learning is further enabled in the CLI module. From the feature representations learned in the instance-MLP, the clustering assignments (i.e., pseudo labels) are generated by self-labeling, which are connected to the clustering assignments learned in the cluster-MLP by minimizing a cross-entropy loss between them, which in fact constrains the consistency between the two levels of contrastive learning. It is noteworthy that the overall network architecture can be trained in an unsupervised manner. Extensive experiments are conducted on six challenging image datasets, which demonstrate the effectiveness of our PICI approach.

For clarity, the main contributions of this work are summarized below.
\begin{itemize}
	\item To enjoy the advantages of masked image modeling, we jointly enforce two types of partial information discrimination learning, i.e., the PISD for masked reconstruction and the PICD for masked contrastive learning, which for the first time, to our knowledge, bridges the gap between the masked image modeling and the deep clustering.
	\item To enhance the consistency between the instance-level contrastive learning and the cluster-level contrastive learning, we design a cross-level interaction mechanism to adaptively guide them in the label space.
	\item A novel deep image clustering approach termed PICI is proposed. Experimental results on a variety of benchmark datasets have confirmed its superiority over the state-of-the-art deep clustering approaches.
\end{itemize}

The remainder of this paper is organized as follows. The related works on self-supervised learning and deep learning are reviewed in Section~\ref{sec:related work}. We present the proposed PID and CLI in Section~\ref{sec:proposed framework}. The quantitative and qualitative experiments are reported in Section~\ref{sec:experiments}, where demonstrate the effectiveness and the superiority of our PICI. Finally, a conclusion of this paper is presented in Section~\ref{sec:conclusion}.

\section{Related Work}
\label{sec:related work}
In this section, we review the related works on self-supervised learning and deep clustering in Sections~\ref{sec:related_work_SSL} and \ref{sec:related_work_DC}, respectively.

\subsection{Self-Supervised Learning}
\label{sec:related_work_SSL}
Recent developments in contrastive learning have greatly advanced the research in self-supervised learning. In a nutshell, the core idea of contrastive learning is to first construct positive and negative pairs, and then to pull the positive pairs close while pushing the negatives far away in the embedding subspace. Previous contrastive learning methods \cite{wu2018unsupervised,chen2020simple,he2020momentum} often require a relatively large number of negative samples for instance discrimination, which in fact treat each sample as a category and may lead to extra computational costs and increased memory storage. For example, the Instance Recognition (IR) method \cite{wu2018unsupervised} utilizes a discrete memory bank to store the features of each sample. Meanwhile, the SimCLR method \cite{chen2020simple} requires a large batch size (e.g., 4096) while the MoCo method \cite{he2020momentum} employs a memory queue to temporarily save the representations produced by a momentum encoder with an exponential moving average.

Some recent studies suggest that the negative pairs are not essential for achieving instance discrimination in contrastive learning \cite{grill2020bootstrap,chen2021exploring,caron2020unsupervised}. For example, both BYOL \cite{grill2020bootstrap} and SimSiam \cite{chen2021exploring} adopt an online predictor to avoid collapsing solutions while eliminating the dependence on negative pairs. Alternatively, the SwAV method \cite{caron2020unsupervised} swaps the predictions where it predicts the code of a view from the representation of another view, which incorporates the clustering into a siamese network.
More recently, the Masked Image Modeling (MIM) \cite{he2022masked,assran2022masked,huang2022contrastive,chen2022context} has emerged as a new trend in self-supervised learning, which designs a prediction task that reconstructs masked images to enhance the representation learning capability of the ViT \cite{DosovitskiyB0WZ21}. Following the emergence of the MAE \cite{he2022masked}, which is a representative model in MIM, the Context Auto-Encoder (CAE) \cite{chen2022context} further incorporates a latent contextual regressor with the alignment constraint, while the Masked Siamese Network (MSN) \cite{assran2022masked} and the Contrastive Masked Auto-Encoder (CMAE) \cite{huang2022contrastive} incorporate the contrastive learning paradigm into the MIM framework.

\subsection{Deep Clustering}
\label{sec:related_work_DC}

In recent years, the deep clustering methods \cite{xie2016unsupervised,guo2017improved,guo2019adaptive,tao2021clustering,li2021contrastive,wu2022deep,deng2023heterogeneous,caron2018deep,zhong2021graph,dang2021doubly,asano2019self} have attracted increasing attention and made significant progress with the support of deep neural networks. Some early works in deep clustering, such the Auto-encoders such as Deep Embedding Clustering (DEC) \cite{xie2016unsupervised}, the Improved Deep Embedding Clustering (IDEC) \cite{guo2017improved}, and the Adaptive Self-Paced Deep Clustering with Data Augmentation (ASPC-DA) \cite{guo2019adaptive}, often utilize the reconstruction loss to pre-train the network and further achieve the clustering result either by associating some clustering loss with a specific layer in the network or by simply performing the $K$-means clustering on the learned representation. Alternatively, the Deep Clustering and Visualization (DCV) method \cite{wu2022deep} integrates the clustering task with data visualization to preserve the geometric structure. The DeepCluster method \cite{caron2018deep} utilizes the self-labeling to enhance the clustering performance by transforming the unsupervised image clustering problem into a supervised one guided by the pseudo-labels.

More recently, the contrastive learning has demonstrated its promising potential in deep clustering. To incorporate the contrastive learning paradigm into the deep clustering, the data augmentation is required to construct the sample pairs, where the samples augmented from the same instance are treated positive pairs, while the others negative pairs. Specifically, the Instance Discrimination and Feature Decorrelation (IDFD) method \cite{tao2021clustering} aims to learn similarities by the instance-level contrastive learning while reducing the correlations within features in the meantime. Furthermore, some recent methods \cite{li2021contrastive,zhong2021graph,dang2021doubly,deng2023heterogeneous} seek to conduct the contrastive learning at both the instance-level and the cluster-level for simultaneous representation learning and clustering, among which a representative method is the Contrastive Clustering (CC) method \cite{li2021contrastive}. Different from the CC method which utilizes two identical and weight-sharing networks for the two augmented views, respectively, Deng et al. \cite{deng2023heterogeneous} presented a Heterogeneous Tri-stream Clustering Network (HTCN), which extends the two-stream contrastive learning network into three streams of heterogeneous networks, including two online networks and a target network, for learning clustering-friendly representations for the image clustering task. 

\section{Proposed Framework}
\label{sec:proposed framework}
In this paper, we present a novel unsupervised deep image clustering model termed PICI, which can be trained in an unsupervised manner. The training of our PICI model mainly involves three learning modules, namely, the PISD model for the self-discrimination learning of the masked images, the PICD module with two levels of contrastive learning, and the CLI module for imposing the cross-level consistency.

Specifically, given an input image, we first perform two types of augmentations on the image and thus obtain two augmented samples. These two augmentations form the two paralled views of the backbone network, which is a Transformer-based encoder. Note that each of the two augmented samples is split into a sequence of patches, which are randomly masked and then fed into the backbone to produce the class tokens [CLS]. Thereafter, three learning modules are jointly incorporated. In the PISD module, a decoder is leveraged to recover the original masked images. In the PICD module, utilized the [CLS] tokens to achieve the instance-level and the cluster-level contrastive learning, which are associated with an instance contrastive loss and a cluster contrastive loss, respectively. Furthermore, the CLI module is exploited to enforce the consistency between the two-levels of contrastive learning via the instance-MLP and the cluster-MLP, respectively, where the set of pseudo labels in the instance-level space are generated by self labeling and the cross-level consistency is imposed through the maximum match cluster labels between the pseudo labels and the cluster assignment from the cluster-level space. Finally, with the overall network trained, the clustering assignments in the cluster-MLP can be obtained as the final clustering result. 


\subsection{Backbone with Parallel Views}
\label{sec:backbone with parallel views}
In the proposed PICI model, we construct two parallel views through data augmentations, which will be fed to Transformer-based encoder and further leveraged for the contrastive learning process. Specifically, we adopt two \textit{different} transformation families for the two views, where the first view is associated with a conventional weak augmentation (without severe distortions) and the second view is associated with a stronger augmentation (with more severe distortions) so as learn more discriminative features from the image. Formally, let the weak augmentation be denoted as $T_{w}$ and the strong augmentation be denoted as $T_{s}$. Given an input image, say, $x_i$, a weak augmentation and a strong one are performed on the image, respectively, leading to its two corresponding views, denoted as $x^a_i=T_{w}(x_i)$ and $x^b_i=T_{s}(x_i)$.

It is noteworthy that most previous contrastive learning works tend to adopt some data augmentations without severe distortions, which are called weak augmentations. In some recent studies, it has been proven that mixing weak and strong augmentations may lead to better contrastive learning performance \cite{wang2022contrastive,deng2023strongly}. In this work, we adopt the weak transformations in the weakly augmented view and the strong transformations in the strongly augmented view.
Given the parallel views, a weight-sharing backbone is used to extract features $h$ (i.e., [CLS]) from the augmented samples. Note that the augmented samples in the two parallel views will be split into sequences of patches and then randomly masked. Thereafter, we adopt the ViT \cite{DosovitskiyB0WZ21} as the backbone, where the self-attention mechanism of the Transformer \cite{vaswani2017attention} can bring in the information of global dependencies for enhanced representation learning.

\subsection{Partial Information Discrimination}
Different from conventional deep clustering methods that mostly work at the full-image scale, in this paper, we seek to enforce two types of partial information discrimination, corresponding to the PISD module and the PICD module, respectively, for representation learning via the masked images. Specifically, the PISD module is utilized to train the network by minimizing the discrimination between the original image and the image recovered from the masked image, which will be described in Section~\ref{sec:partial information reconstruction}. Meanwhile, the PICD module is incorporated to perform two levels of contrastive learning, aiming to minimize the discrimination between the  positive pairs while maximizing that between the negative pairs at the instance-level and the cluster-level, respectively, which will be described in Section~\ref{sec:partial information contrastiveness}.

\subsubsection{Partial Information Self-Discrimination (PISD)}
\label{sec:partial information reconstruction}
For an augmented image, we divide it into a sequence of regular non-overlapping patches following the ViT \cite{DosovitskiyB0WZ21}. Then a subset of patches are randomly selected as visible patches and the rest of them as masked (or invisible) patches. Following the MAE \cite{he2022masked}, we select the random patches to be masked w.r.t. a uniform distribution. It is argued that the masked random sampling with a certain masking ratio (e.g., 50\%) largely eliminates the redundancy of image data \cite{he2022masked} and the sparse (masked) input is conducive to improving the ability of representation learning for images.

Formally, let the Transformer encoder be denoted as $F(\cdot)$, which operates only on the unmasked patches. These unmasked (or visible) patches are projected with position embeddings, which are employed to retain the positional information for the later recovery. Concretely, we adopt the standard learnable 1-D position embeddings and utilize the resultant sequence of embedding vectors as the input to the encoder. Note that the masked patches are removed and only the unmasked tokens are used, which only takes partial information into consideration and also alleviates the time and memory consumption of the model training process.

In the PISD module, to learn the discriminative representations by recovering the missing patches, we incorporate a decoder $D(\cdot)$ through another series of Transformer blocks for image reconstruction. Specifically, to reconstruct the image from the embeddings produced by the encoder $F(\cdot)$, we add positional embeddings to all tokens so that the masked tokens will be located in their corresponding positions in the image. With the encoder and the decoder forming an auto-encoder, it is employed to reconstruct the input image by predicting the pixel values for all masked patches. Without loss of generality, we calculate the Mean Squared Error (MSE) between the original and reconstructed images at the pixel level as the reconstruction loss. Then we have the overall reconstruction loss of the PISD module as follows:
\begin{equation}
	\mathcal{L}_{PISD} = \frac{1}{2} (\ell_{re}^a + \ell_{re}^b) .
	\label{equ:pir}
\end{equation}
where $\ell_{re}^a$ and $\ell_{re}^b$  denote the MSE losses for the first and second views, respectively.

\subsubsection{Partial Information Contrastive Discrimination (PICD)}
\label{sec:partial information contrastiveness}
In the PICD module, to simultaneously enforce the instance-level and cluster-level contrastive learning, two independent MLP projectors are utilized to project the representations (i.e., [CLS]) extracted by the backbone to the instance-level and cluster-level subspaces, respectively, which will further be associated with the instance loss and the cluster loss for contrastive representation learning. Before delving into the details of these two contrastive losses, we first present the overall loss of the PICD module, that is
\begin{equation}
	\mathcal{L}_{PICD} = \mathcal{L}_{ins} + \mathcal{L}_{clu}.
	\label{equ:pic}
\end{equation}
where $\mathcal{L}_{ins}$ is the instance contrastive loss associated with the instance-MLP, and $\mathcal{L}_{clu}$ is the cluster contrastive loss associated with the cluster-MLP.

\paragraph{Instance-level Contrastiveness}
\label{sec:instance-wise contrastiveness}
The instance-level contrastive learning essentially works by maximizing the similarities between the positive sample pairs while minimizing that of the negative ones. In contrastive learning, it is a fundamental issue to define the positive and negative pairs. In recent years, many methods for constructing positive and negative sample pairs have been developed. For example, one can define the pairs of within-class samples to be positive and the between-class pairs to be negative. However, without prior knowledge, whether two arbitrary samples belong to the same class is hard to determine. In this paper, following the standard protocol in contrastive learning\cite{chen2020simple, li2021contrastive}, we regard the pair of samples augmented from the same sample as positive and other pairs as negative.

Given a mini-batch of $N$ samples, our PICI performs weak and strong data augmentations on each sample and then $2\cdot N$ augmented samples, denoted as $\{x_1^a, \dots, x_N^a, x_1^b, \dots, x_N^b \}$,
are produced. For a specific sample $x_{i}$, the pair of $\{x_i^a,x_i^b\}$ is treated as its exclusive positive pair while the other $2\cdot (N-1)$ pairs as its negative pairs.
Note that directly conducting contrastive learning on the feature representation $h$ may induce information loss \cite{chen2020simple}. Hence we stack a two-layer nonlinear MLP $G_{I}(\cdot)$, i.e., the instance-MLP, to map the feature representations $h^a_i$ and $h^b_i$ onto a low-dimensional subspace, denoted as $Z^a_i=G_I(h^a_i)$ and $Z^b_i=G_I(h^b_i)$, respectively. 

Let the instance representations set for the first view be denoted as $\emph{I}^a = \{Z_1^a,Z_2^a,\dots,Z_N^a\}$ and that for the second view as $\emph{I}^b$. Then the pair-wise similarity between two feature vectors is measured by the cosine similarity, that is
\begin{equation}
	sim(Z^{k_1}_i, Z^{k_2}_j) = \frac{({Z^{k_1}_i})^{\top}(Z^{k_2}_j)}{\|Z^{k_1}_i\|\cdot\|Z^{k_2}_j\|},
	\label{equ:cosine_sim}
\end{equation}
with $k_1,k_2 \in \{a,b\}$ and $i,j \in[1,N]$. It is noteworthy that if $Z_i$ and $Z_j$ are normalized to unit norm, the cosine similarity in Eq.~(\ref{equ:cosine_sim}) can be simplified into a dot-product form, that is
\begin{equation}
	 sim(Z^{k_1}_i, Z^{k_2}_j) = ({Z^{k_1}_i})^{\top}(Z^{k_2}_j).
\end{equation}

Further, we utilize the InfoNCE loss \cite{oord2018representation} for instance-level contrastive learning. Thus, we have the contrastive loss for a given sample $x^a_i$ as
\begin{equation}
	\begin{aligned}
		\ell^a_i =-log\frac{\exp(sim(Z^a_i,Z^b_i)/\tau_I)}{ {\textstyle \sum_{k\in {a,b}}\sum_{j=1}^{N}}\exp(sim(Z^a_i,Z^k_j)/\tau_I)},
		\label{equ:ins_loss_a}
	\end{aligned}
\end{equation}
where $\tau_I$ is the instance temperature parameter that adjusts the degree of attraction and repellence between samples.

Finally, the instance contrastive loss for a mini-batch of $N$ input images can be represented as
\begin{equation}
	\mathcal{L}_{ins} = \frac{1}{2\cdot N}\sum_{i=1}^{N}{(\ell^a_i + \ell^b_i)}.
	\label{equ:ins_loss}
\end{equation}
where $\ell^a_i$ is the instance contrastive loss for a given sample $x_i$ for the first view and $\ell^b_i$ for the second view.

\paragraph{Cluster-level Contrastiveness}
\label{sec:cluster-wise contrastiveness}
The idea of ``label as representation" in online clustering implies that, when a data sample is projected into a space, where the number of dimensions corresponds to the number of clusters, the $i$-th element of its feature vector can be regarded as its likelihood of being part of the $i$-th cluster. Consequently, this feature vector can be seen as the data sample's soft label \cite{li2022twin}.

Similar to the instance-MLP $G_I(\cdot)$ in the instance-level contrastive learning, we employ another two-layer MLP $G_C(\cdot)$, i.e., the cluster-MLP, with an extra softmax layer to project the representations $h^a_i$ and $h^b_i$ into an $M$-dimensional space. Formally, let $C^a = \{Y_1^a, Y_2^a,\dots,Y_N^a\} \in \mathbb{R}^{N\times M}$ represent the cluster assignment probabilities for a mini-batch of $N$ samples in the first view, and let $C^b = \{Y_1^b, Y_2^b,\dots,Y_N^b\}$ represent those in the second view, where $M$ is the number of clusters. Consequently, the $i$-th column of $C^a$ can be interpreted as a representation of the $i$-th cluster, and all columns should be distinct from one another. Let $\{Y_i^a,Y_i^b\}$ represent a positive cluster pair while leaving the other $2\cdot M -2$ pairs as negative pairs. Here, the cosine similarity is employed to measure the similarity between cluster pairs. 
\begin{equation}
	sim(Y^{k_1}_i, Y^{k_2}_j) = \frac{({Y^{k_1}_i})^{\top} (Y^{k_2}_j)}{\|Y^{k_1}_i\|\cdot\|Y^{k_2}_j\|},
	\label{equ:cosine_sim2}
\end{equation}
with $k_1,k_2 \in \{a,b\}$ and $i,j \in[1,M]$. Here, the cluster-level contrastive loss is defined to differentiate cluster $Y_i^a$ from all other clusters except its counterpart $Y_i^b$. Thus, the contrastive loss for cluster $Y_i^a$ can be calculated as
\begin{equation}
	\begin{aligned}
		\tilde{\ell^a_i} =-log\frac{\exp(sim(Y^a_i,Y^b_i)/\tau_C)}{ {\textstyle \sum_{k\in \{a,b\}}\sum_{j=1}^{M}}\exp(sim(Y^a_i,Y^k_j)/\tau_C)}
		\label{equ:clu_loss_a}
	\end{aligned}
\end{equation}
where $\tau_C$ is the cluster temperature parameter. Directly minimizing the above contrastive loss might lead to a trivial solution that assigns most samples to a single cluster or a few clusters. To circumvent this issue, an entropy term is introduced to constrain the cluster assignment probabilities, that is
\begin{equation}
	\mathcal{H}(C) = -\sum_{i=1}^{M} [P(Y_i^a)\log_{}{P(Y_i^a)} + P(Y_i^b)\log_{}{P(Y_i^b)}]
	\label{equ:entropy}
\end{equation}

\begin{equation}
	P(Y_{i}^{k})=\frac{\sum_{j=1}^{N} C_{j i}^{k}}{\|{Y}_{i}^{k}\|_{1}}, k\in \{a,b\}, i\in \left[1, M\right]
\end{equation}
where $C_{ji}^{k}$ denotes the probability of sample $j$ being assigned to cluster $i$. Therefore, the overall cluster contrastive loss function can be defined as follows:
\begin{equation}
	\mathcal{L}_{clu} = \frac{1}{2\cdot M}\sum_{i=1}^M\left(\tilde{\ell^a_i} + \tilde{\ell^b_i}\right)-\mathcal{H}(C)
	\label{equ:clu_loss}
\end{equation}

\subsection{Cross-level Interaction (CLI)}
\label{sec:cross-level interaction}
In the PICD module, as described in Section~\ref{sec:partial information contrastiveness}, the model simultaneously learns the instance-wise features and the cluster assignments $\{C^k\},k\in \{a,b\}$ through two levels of contrastive learning. These two levels of contrastive learning effectively capture the instance-level and cluster-level contrastive information, respectively. To strengthen the connection between these two learning levels, in this section, we further incorporate the CLI module to achieve the joint learning with the cross-level consistency adaptively enforced.

However, in practice \cite{li2021contrastive, xu2022multi}, the instance-level subspace is semantic-richer compared with the cluster-level subspace. Based on this observation, we propose Cross-level Interaction (CLI) to build a bridge between instance- and cluster-level subspaces. Let $\{C^k\}$ denote anchors and align them with the clusters among $\emph{I}^k$. In this way, the cluster information contained in the instance-level subspace is leveraged to improve the clustering effect of the semantic labels.

Specifically, we employ the $K$-means clustering to produce the pseudo-labels of all samples in the instance-level subspace. For the $k$-th view with $k\in\{a,b\}$, let $\{u^k_m\}^M_{m=1} \in \mathbb{R}^I$ denote the $M$ cluster centroids. Thus, we have
\begin{equation}
	\min_{u^k_1,u^k_2,\dots,u^k_M} \sum_{i=1}^{N} \sum_{j=1}^{M} \left\| Z^k_i-u^k_j\right\|^{2}_{2}.
	\label{equ:kmeans_centroids}
\end{equation}
Then the pseudo-labels of all samples $\boldsymbol p^k \in \mathbb{R}^N$ (for $k\in\{a,b\}$) are obtained as
\begin{equation}
	p^k_i = \mathop{\arg\min}_{j} \left\| Z^k_i-u^k_j\right\|^{2}_{2}.
	\label{equ:pseudo_labels}
\end{equation}
Let $\boldsymbol{q}^k \in \mathbb{R}^N$ (for $k\in\{a,b\}$) denote the cluster labels generated from the cluster-level subspace, which can be represented as
\begin{equation}
	q^k_i = \underset{j}{\arg\max} q^k_{ij}.
\end{equation} 
Further, to enable the interation between the instance-level and cluster-level, we treat $\boldsymbol q^k$ as the anchors to align with $\boldsymbol p^k$ according to the maximum matching criterion \cite{xu2022multi}.
In practice, the problem of maximum matching cluster labels can be solved by the Hungarian algorithm \cite{jonker1986improving}. Thereafter, the modified cluster assignments $\tilde{p}_i^k$ are generated, where $\tilde{p}_i^k \in \{0,1\}^{M} $ for the $i$-th sample is defined as a one-hot vector. The $m$-th element of $\tilde{p}_i^k $ is 1 if there exists $s \in \{1,2,\dots,M\}$ such that $\mathbbm{1}[{w}_{ms}^k=1] \cdot \mathbbm{1}[p_i^k=s]=1$. Here, $ \mathbbm{1}[\cdot]$ is the indicator function.

Formally, we adopt the cross-entropy loss to align the distribution of the pseudo-labels with the cluster assignments, that is
\begin{equation}
	\mathcal{L}_{CLI}=- \frac{1}{2}\cdot{\textstyle \sum_{k\in \{a,b\}}} \tilde{P}^k \log C^k,
	\label{equ:ci}
\end{equation}
where ${\tilde{P}}^k = \{\tilde{p}_1^k,\tilde{p}_2^k,\dots,\tilde{p}_N^k\}\in \mathbb{R}^{N\times M}$ and ${C}^k = \{q_1^k,q_2^k,\dots,q_N^k\}\in \mathbb{R}^{N\times M}$.
In this way, the semantic-rich instance features are leveraged to guide the cluster-level learning. This, in turn, optimizes the backbone network via back-propagation, which subsequently benefits the instance-level learning. Thereby, the overall representation learning of the proposed model is enhanced through the adaptive interaction between the two levels of contrastive learning in the CLI module.

\subsection{Training Strategy}
\label{sec:training strategy}

In this work, the network training is performed in three stages. In the first stage, to enhance the stability and representation learning capability of the proposed model, we first pre-train the masked auto-encoder with the reconstruction loss as defined in Eq.~(\ref{equ:pir}). Then, the reconstruction-based learning in PISD and the two levels of contrastive learning in PICD are jointly enforced in the second stage. Finally, the contrastive learning in PICD and the cross-level learning in CLI are further conducted in the boosting stage. For clarity, the overall process of the proposed PICI approach is summarized in Algorithm~\ref{algorithm}.

\begin{algorithm}[!h]
	\caption{Algorithm for PICI}
	\label{algorithm}
	\KwIn{Dataset $\mathcal{X}$; Pre-training epochs $E_1$; Training epochs $E_2$; Boosting epochs $E_3$; Batch size $N$; Masked random sampling $\mathbb{M}$; Temperature parameters $\tau_I$ and $\tau_C$; Cluster number $M$; Network structure of $\mathcal{T}$, $F$, $D$, $G_I$, and $G_C$.}
	
	\KwOut{Cluster assignments.}
	
	\tcp{Pre-training}
	\For{epoch = 1 to $E_1$}{
		Sample a mini-batch $\{x_i\}_{i=1}^N$ from $\mathcal{X}$\\
		Sample two augmentations $T_w, T_s \sim \mathcal{T}$\\
		$\tilde{h}^a_i = F(\mathbb{M}(T_w(x_i)))$, $\tilde{h}^b_i = F(\mathbb{M}(T_s(x_i)))$,
		$\tilde{x}^a_i = D(\tilde{h}^a_i)$, $\tilde{x}^b_i = D(\tilde{h}^b_i)$\\
		Compute $\mathcal{L}_{PISD}$ through Eq.~(\ref{equ:pir})\\
		Update $F, D$ to minimize $\mathcal{L}_{PISD}$
	}
	\tcp{Training}
	\For{epoch = 1 to $E_2$}{
		Conduct dual contrastive learning\\
		Compute $\mathcal{L}_{PICD}$ through Eq.~(\ref{equ:pic})\\
		Update $F, D, G_I, G_C$ to minimize $\mathcal{L}_{PISD} + \mathcal{L}_{PICD}$
	}
	\tcp{Boosting}
	\For{epoch = 1 to $E_3$}{
		Match cluster labels between instance-level and cluster-level spaces by solving Eq.~(\ref{equ:assign}).\\
		Compute $\mathcal{L}_{CLI}$ through Eq.~(\ref{equ:ci})\\
		Update $F, G_I, G_C$ to minimize $\mathcal{L}_{PICD} + \mathcal{L}_{CLI}$
	}
	\tcp{Test}
	\For{$x$ in $\mathcal{X}$}{
		Extract representations by $h = F(x)$
		Calculate the cluster assignment by $c=\arg\max G_C(h)$
	}
\end{algorithm}

\section{Experiments}
\label{sec:experiments}
In this section, we conduct extensive experiments to benchmark the proposed PICI approach against a variety of non-deep and deep clustering approaches on six real-world image datasets. In addition, we present qualitative analyses and ablation experiments to provide a more comprehensive and clear perspective on our proposed approach.

\subsection{Implementation Details}
\label{sec:implementation details}

In this work, we adopt two distinct families of data augmentations to generate the weak and strong augmentations, respectively. Specifically, the strong augmentation $T_s$ is randomly drawn from the augmentation family that includes ResizedCrop, ColorJitter, Grayscale, HorizontalFlip, and GaussianBlur, while the weak augmentation $T_w$ merely resizes and normalizes the images. A lightweight version of ViT, namely the ViT-Small \cite{touvron2021training}, is utilized as the backbone, with a dimension size of 384 and 6 blocks of encoder layers. In contrast, the corresponding decoder consists of 8 Transformer blocks and 16 multi-heads with a dimension of 512. The combined encoder and decoder can be regarded as a vanilla MAE \cite{he2022masked}. Each input image is resized to dimensions of $224 \times 224$.

In PICD, two types of projectors are incorporated, including the instance-MLP with an output dimension of 128 and the cluster-MLP with the output dimension setting to the target cluster number. The temperature parameters $\tau_I$ and $\tau_C$ are set to 0.5 and 1.0, respectively.  To optimize the network, we use the Adam optimizer with an initial learning rate $1e^{- 4}$. 
Note that the masked random sampling varies between the training phase and the testing phase. During training, the masking ratio is set to $50\%$ across all datasets. During testing, the masked random sampling is omitted. The model is pre-trained for 200 epochs and trained for 800 epochs, followed by 50 boosting epochs for all image datasets. The batch size is set to 96. All experiments are executed on a single NVIDIA RTX 3090 GPU, using the Ubuntu 18.04 platform with CUDA 11.0 and Pytorch 1.7.0.

\subsection{Datasets and Evaluation Metrics}
In our experiments, six real-world image datasets are used for evaluation, which are described as follows:
\begin{itemize}
	\item \textbf{RSOD} \cite{long2017accurate} is an open remote sensing dataset, which consists a total of 976 images and 4 classes, namely, the aircraft, the playground, the overpass, and the oil tank.
	\item \textbf{UC-Merced} \cite{yang2010bag} is a land use image dataset with 21 classes. Each class includes 100 images. The pixel-size of each image is 256$\times$256.
	\item \textbf{SIRI-WHU} \cite{zhao2015dirichlet} is a 12-class remote sensing image dataset, which is constructed by the RS\_IDEA Group in Wuhan University (SIRI-WHU). Each class in the dataset includes 200 images with a size of 200$\times$200.
	\item \textbf{AID} \cite{xia2017aid} is a large-scale remote sensing dataset, which includes 30 aerial scene types and a total of 10,000 images.
	\item \textbf{D0} \cite{xie2018multi} is an image dataset with 40 common pest species, which consists of a total of 4,508 images.
	\item \textbf{Chaoyang} \cite{zhu2021hard} comprises 6,160 images from 4 classes of colon slides, collected from Chaoyang Hospital.
\end{itemize}

For clarity, we summarize the statistics of the six image datasets in Table~\ref{table-dataset}. Note that the images in the RSOD dataset are not uniform in size. Following the standard evaluation protocol for the image clustering task, we adopt three widely-used evaluation metrics in our experiments, including the Normalized Mutual Information (NMI) \cite{Choudhury21_tetci}, the accuracy (ACC) \cite{Huang2020}, and the Adjusted Rand Index (ARI) \cite{Huang2021}. The value range for NMI and ACC is $[0,1]$, whereas ARI varies in the range of $[-1,1]$. It's worth noting that higher values of these metrics indicate better clustering results.

\begin{table}[!t]
	\caption{The image datasets used in our experiments.}
	\label{table-dataset}
	\centering
	\renewcommand{\arraystretch}{1.25}
	\setlength{\tabcolsep}{4.2mm}{
		\begin{tabular}{l c c c}
			\toprule
			Dataset   & \#Samples & \#Classes & \#Image Size\\
			\midrule
			RSOD \cite{long2017accurate}  & 976  & 4 &--\\
			UC-Merced \cite{yang2010bag}  & 2,100  & 21 &256$\times$256\\
			SIRI-WHU \cite{zhao2015dirichlet}  & 2,400  & 12 &200$\times$200\\
			AID \cite{xia2017aid}  & 10,000  & 30 &600$\times$600\\
			D0 \cite{xie2018multi}      & 4,508 & 40 &200$\times$200\\
			Chaoyang \cite{zhu2021hard}     & 6,160 & 4 &512$\times$512\\
			\bottomrule
	\end{tabular}}
\end{table}

\begin{figure}[!htb]
	\centering
	\subfigure[RSOD]{
		\includegraphics[width=0.486\textwidth]{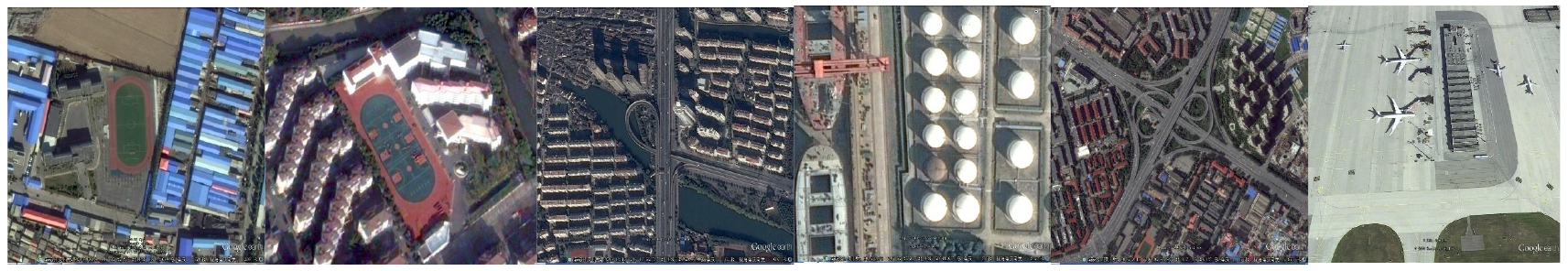}}
	\subfigure[UC-Merced]{
		\includegraphics[width=0.486\textwidth]{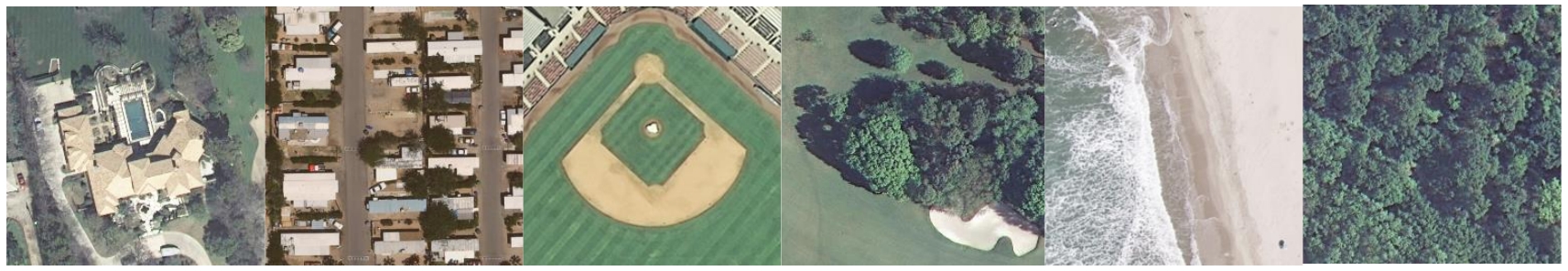}}\vskip -0.015in
	\subfigure[SIRI-WHU]{
		\includegraphics[width=0.486\textwidth]{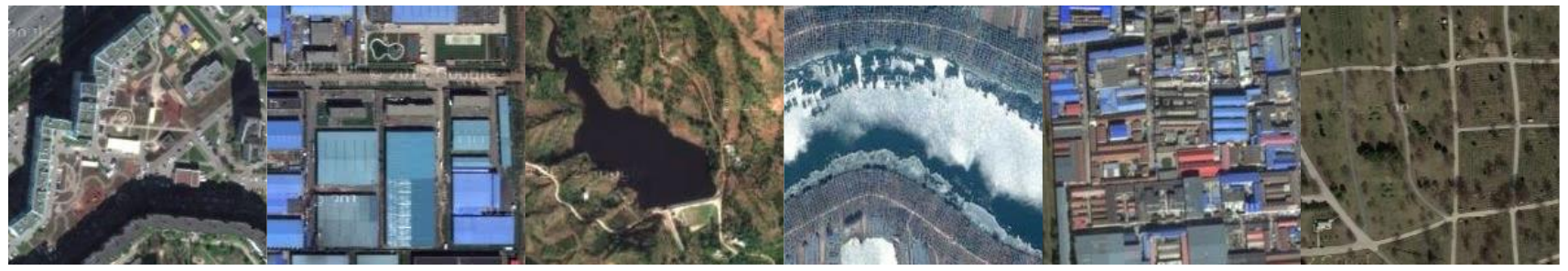}}
	\subfigure[AID]{
		\includegraphics[width=0.486\textwidth]{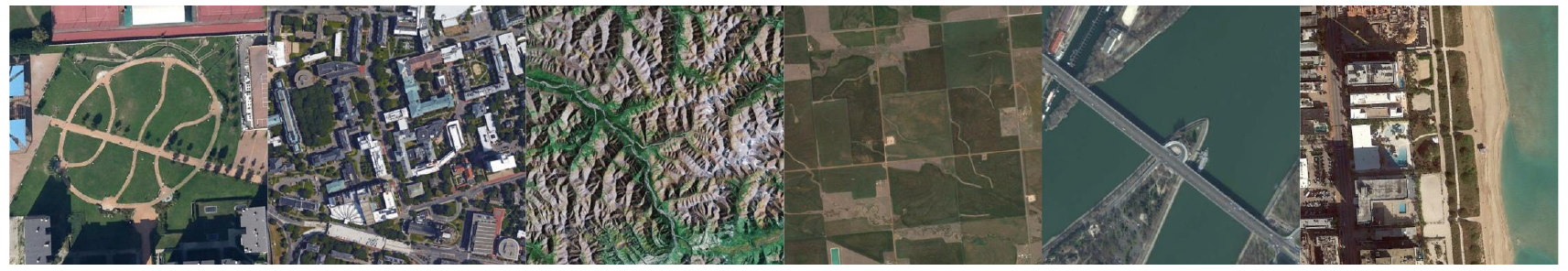}}\vskip -0.015in
	\subfigure[D0-40]{
		\includegraphics[width=0.486\textwidth]{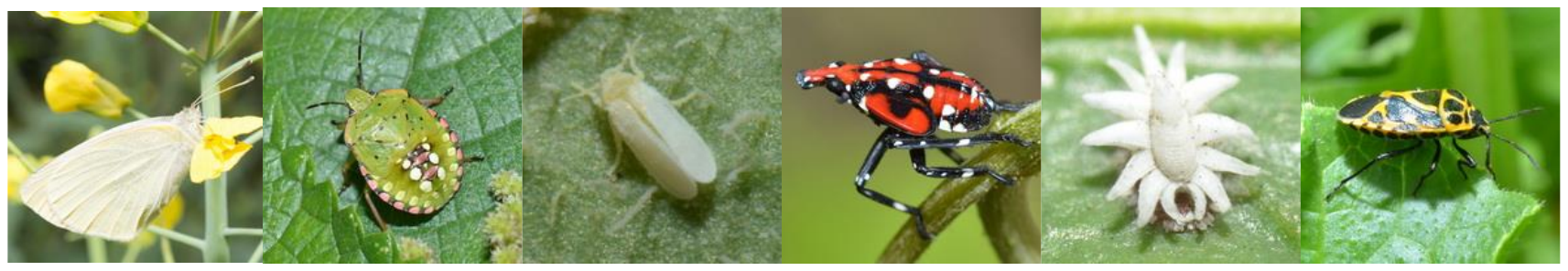}}
	\subfigure[Chaoyang]{
		\includegraphics[width=0.486\textwidth]{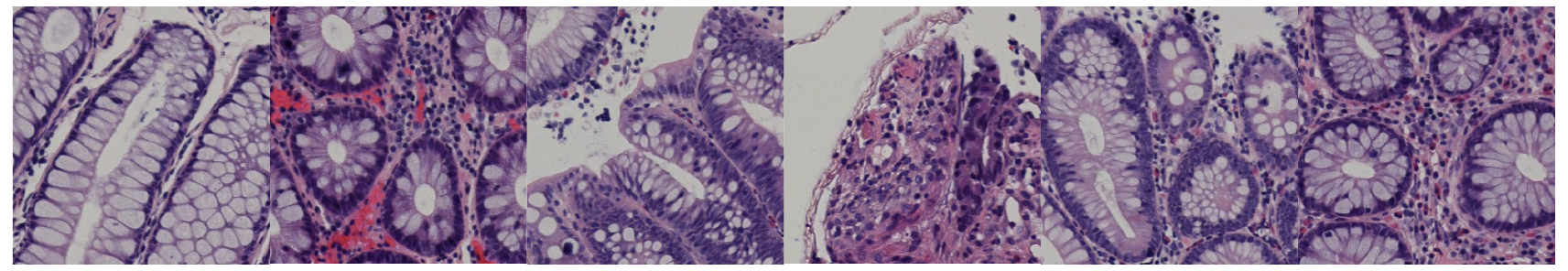}}\vskip -0.015in
	
	\caption{Some examples of the six image datasets used for evaluation, including four remote sensing datasets \cite{yang2010bag,long2017accurate,zhao2015dirichlet,xia2017aid}, a crop pest dataset \cite{xie2018multi}, and a medical dataset \cite{zhu2021hard}.}
	\label{fig:datasets_examples}
\end{figure}

\subsection{Results and Analysis}

\begin{table*}[!t]\vskip 0.1in
	\renewcommand\arraystretch{1}
	\centering
	\small
	\setlength{\tabcolsep}{1.2mm}
	\caption{The \textbf{NMI} scores of different image clustering methods on six datasets.}
	\label{tab-NMI}
	\begin{threeparttable}
		\begin{tabular}{m{2.1cm}<{}|m{1.4cm}<{\centering}m{1.4cm}<{\centering}m{1.4cm}<{\centering}m{1.4cm}<{\centering}m{1.4cm}<{\centering}m{1.4cm}<{\centering}}
			\toprule
			Dataset &RSOD  &UC-Merced &SIRI-WHU &AID   &D0-40   &Chaoyang \\
			\midrule
			$K$-means \cite{macqueen1967some}	&0.162	&0.204	&0.145	&0.209	&0.299	&0.024\\
			SC \cite{zelnik2005self}	&0.146	&0.211	&0.161	&0.189	&0.305		&0.022\\
			AC \cite{gowda1978agglomerative}	&0.168	&0.214	&0.166	&0.204	&0.319	&0.026\\
			NMF \cite{cai2009locality}	&0.176	&0.202	&0.245	&0.193	&0.255	&0.018\\
			PCA \cite{martinez01_pca}	&0.163	&0.206	&0.164	&0.216	&0.308	&0.024\\
			BIRCH \cite{Zhang1996}	&0.148	&0.225	&0.162	&0.204	&0.315	&0.026\\
			GMM	\cite{fraley03} &0.160	&0.198	&0.160	&0.205	&0.289	&0.024\\
			\midrule
			DEC	\cite{xie2016unsupervised} &0.296	&0.120	&0.183	&0.217	&0.328	&0.001\\
			IDEC \cite{guo2017improved}	&0.209	&0.119	&0.178	&0.207	&0.309	&0.001\\
			ASPC-DA \cite{guo2019adaptive}	&0.054	&0.137	&0.103	&0.060	&0.153	&0.026\\
			IDFD \cite{tao2021clustering}	&0.391	&0.572	&0.540	&0.696	&0.663	&0.309\\
			CC \cite{li2021contrastive}	&0.457	&0.609	&0.603	&0.752	&0.693	&0.365\\
			DCV \cite{wu2022deep} &0.178 &0.102 &0.128 &0.127 &0.139 &0.024  \\
			HTCN \cite{deng2023heterogeneous} &0.557 &0.596 &0.534 &0.797 &0.721 &0.276\\
			PICI (Ours)  &\textbf{0.583} &\textbf{0.681}	&\textbf{0.658}	&\textbf{0.800} &\textbf{0.731} &\textbf{0.382} \\
			\bottomrule
		\end{tabular}
	\end{threeparttable}
\end{table*}

\begin{table*}[!t]
	\renewcommand\arraystretch{1}
	\centering
	\small
	\setlength{\tabcolsep}{1.2mm}
	\caption{The \textbf{ACC} scores of different image clustering methods on six datasets.}
	\label{tab-ACC}
	\begin{threeparttable} \begin{tabular}{m{2.1cm}<{}|m{1.4cm}<{\centering}m{1.4cm}<{\centering}m{1.4cm}<{\centering}m{1.4cm}<{\centering}m{1.4cm}<{\centering}m{1.4cm}<{\centering}}
			\toprule
			Dataset &RSOD  &UC-Merced &SIRI-WHU &AID   &D0-40   &Chaoyang\\
			\midrule
			$K$-means \cite{macqueen1967some}	&0.388	&0.200	&0.229	&0.163	&0.204	&0.320\\
			SC \cite{zelnik2005self}	&0.425	&0.183	&0.210	&0.123	&0.195	&0.312\\
			AC \cite{gowda1978agglomerative}	&0.371	&0.188	&0.222	&0.151	&0.209	&0.329\\
			NMF	\cite{cai2009locality} &0.420	&0.208	&0.275	&0.161	&0.187	&0.305\\
			PCA \cite{martinez01_pca}	&0.388	&0.198	&0.227	&0.173	&0.220	&0.320\\
			BIRCH \cite{Zhang1996}	&0.396	&0.202	&0.222	&0.147	&0.205		&0.329\\
			GMM	\cite{fraley03} &0.382	&0.193	&0.239	&0.169	&0.189	&0.318	\\
			\midrule
			DEC	\cite{xie2016unsupervised} &0.534	&0.147	&0.257	&0.185	&0.232	&0.421\\
			IDEC \cite{guo2017improved}	&0.458	&0.141	&0.255	&0.192	&0.213	&0.424\\
			ASPC-DA \cite{guo2019adaptive}	&0.464	&0.073	&0.183	&0.079	&0.107	&0.325\\
			IDFD \cite{tao2021clustering}	&0.595	&0.456	&0.545	&0.628	&0.507	&0.512\\
			CC \cite{li2021contrastive}	&0.538	&0.480	&0.604	&0.622	&0.511	&0.575\\
			DCV \cite{wu2022deep} &0.418 &0.121 &0.195 &0.100 &0.095 &0.321 \\
			HTCN \cite{deng2023heterogeneous} &0.584 &0.508 &0.496 &0.709 &\textbf{0.576} &0.547 \\
			PICI (Ours)  &\textbf{0.772} &\textbf{0.634} &\textbf{0.672} &\textbf{0.748} &0.568	&\textbf{0.595}\\
			\bottomrule
		\end{tabular}
	\end{threeparttable}
\end{table*}

\begin{table*}[!t]
	\renewcommand\arraystretch{1}
	\centering
	\small
	\setlength{\tabcolsep}{1.2mm}
	\caption{The \textbf{ARI} scores of different image clustering methods on six datasets.}
	\label{tab-ARI}
	\begin{threeparttable} \begin{tabular}{m{2.1cm}<{}|m{1.4cm}<{\centering}m{1.4cm}<{\centering}m{1.4cm}<{\centering}m{1.4cm}<{\centering}m{1.4cm}<{\centering}m{1.4cm}<{\centering}}
			\toprule
			Dataset &RSOD  &UC-Merced &SIRI-WHU &AID   &D0-40  &Chaoyang \\
			\midrule
			$K$-means \cite{macqueen1967some}	&0.075	&0.065	&0.053	&0.051	&0.080	&0.017\\
			SC \cite{zelnik2005self}	&0.096	&0.038	&0.041	&0.029	&0.039	&0.005\\
			AC \cite{gowda1978agglomerative}	&0.071	&0.057	&0.057	&0.048	&0.080	&0.010\\
			NMF \cite{cai2009locality}	&0.052	&0.089	&0.118	&0.056	&0.068	&0.002\\
			PCA \cite{martinez01_pca}	&0.075	&0.064	&0.063	&0.054	&0.088	&0.017\\
			BIRCH \cite{Zhang1996}	&0.068	&0.066	&0.049	&0.046	&0.080		&0.010\\
			GMM	\cite{fraley03} &0.069	&0.062	&0.062	&0.053	&0.074	&0.016	\\
			\midrule
			DEC	\cite{xie2016unsupervised} &0.325	&0.053	&0.083	&0.075	&0.105	&0.006\\
			IDEC \cite{guo2017improved}	&0.144	&0.042	&0.079	&0.073	&0.093	&--	\\
			ASPC-DA \cite{guo2019adaptive}	&0.005	&0.002	&0.035	&0.014	&0.021	&0.005\\
			IDFD \cite{tao2021clustering}	&0.362	&0.354	&0.389	&0.547	&0.439	&0.259\\
			CC \cite{li2021contrastive}	&0.371	&0.356	&0.450	&0.550	&0.423	&0.343\\
			DCV \cite{wu2022deep} &0.144 &0.020 &0.043 &0.025 &0.017 &0.017  \\
			HTCN \cite{deng2023heterogeneous} &0.465 &0.359 &0.336 &0.646 &0.470 &0.262\\
			PICI (Ours)  &\textbf{0.510} &\textbf{0.492}	&\textbf{0.518}	&\textbf{0.663} &\textbf{0.500} &\textbf{0.382}\\
			\bottomrule
		\end{tabular}
	\end{threeparttable}\vskip 0.05in
\end{table*}

In this section, we evluate the proposed PICI method against several state-of-the-art clustering methods on six real-world image datasets. These methods can be divided into two categories, i.e., the non-deep clustering methods and the deep clustering methods. The non-deep clustering methods include $K$-means \cite{macqueen1967some}, Spectral Clustering (SC) \cite{zelnik2005self}, Agglomerative Clustering (AC) \cite{gowda1978agglomerative}, Non-negative Matrix Factorization (NMF) \cite{cai2009locality}, Principle Component Analysis (PCA) \cite{martinez01_pca}, Balanced Iterative Reducing and Clustering using Hierarchies (BIRCH) \cite{Zhang1996} and GMM \cite{fraley03}. The deep clustering methods include Deep Embedding Clustering (DEC) \cite{xie2016unsupervised}, Improved Deep Embedding Clustering (IDEC) \cite{guo2017improved}, Adaptive Self-Paced Deep Clustering with Data Augmentation (ASPC-DA) \cite{guo2019adaptive}, Instance Discrimination and Feature Decorrelation (IDFD) \cite{tao2021clustering}, Contrastive Clustering (CC) \cite{li2021contrastive}, and Heterogeneous Tri-stream Clustering Network (HTCN) \cite{deng2023heterogeneous}. For NMF and PCA, the clustering results are obtained by performing $K$-means on the extracted features. For the other algorithms, the model settings will be set as suggested by their corresponding papers. 

The experimental results w.r.t. NMI, ACC, and ARI of different non-deep and deep clustering methods are reported in Tables~\ref{tab-NMI}, \ref{tab-ACC}, and \ref{tab-ARI}, respectively.
It is obvious that the deep clustering methods perform much more effectively than the non-deep ones on the image datasets, probably due to the robust representation learning ability of deep neural networks. According to the results shown in Tables~\ref{tab-NMI} and \ref{tab-ARI}, our proposed PICI method outperforms the baseline methods on all the six benchmark datasets. Especially, PICI surpasses the most competitive baseline (i.e., IDFD) by 0.177 on RSOD in terms of ACC. On the UC-Merced dataset, our model results in an NMI of 0.681, an ACC of 0.634, and an ARI of 0.492, which exhibit a significant margin of 11.8\%, 24.8\%, and 36.7\%, respectively, over the second bast scores. On the other datasets, similar advantages of PICI can also be seen in comparison with the other non-deep and deep clustering methods. The remarkable performance of the proposed PICI framework demonstrates its robust representation learning and clustering ability.

\subsection{Ablation Study}
In this section, we experimentally analyze the influence of different components in PICI. Specifically, the influence of the ViT architecture in PISD, the influence of the PISD and CLI modules, and the influence of the two contrastive projectors in PICD will be tested in Sections.

\begin{table}[!t]\scriptsize
	\caption{The influence of ViT architecture.}
	\label{tab:ablation_backbone}
	\renewcommand
	\arraystretch{1.25}
	\centering
	\scalebox{1}{
		\begin{threeparttable}
			\begin{tabular}{m{1.4cm}<{\centering}|m{1cm}<{\centering}m{1cm}<{\centering}m{1cm}<{\centering}m{1cm}<{\centering}| m{0.6cm}<{\centering}m{0.6cm}<{\centering}m{0.6cm}<{\centering}}
				\toprule
				Dataset &Model  &Dimension &\#Layers &\#Heads &NMI &ACC &ARI\\
				\midrule
				\multirow{3}*{RSOD}
				&\mbox{ViT-Tiny} &192 &4 &12 &0.523 &0.750 &0.480\\
				&\mbox{ViT-Small} &384 &6 &12 &\textbf{0.583} &\textbf{0.772} &\textbf{0.510}\\
				&\mbox{ViT-Base} &768 &12 &12 &-- &-- &--\\
				\midrule
				\multirow{3}*{Chaoyang}
				&\mbox{ViT-Tiny} &192 &4 &12 &0.335 &0.572 &0.330\\
				&\mbox{ViT-Small} &384 &6 &12 &\textbf{0.382} &\textbf{0.595} &\textbf{0.382}\\
				&\mbox{ViT-Base} &768 &12 &12 &0.357 &0.579 &0.346\\
				\bottomrule
			\end{tabular}
	\end{threeparttable}}
\end{table}

\subsubsection{Influence of ViT Architecture}
\label{sec:influence_vit}

To assess the impact of the backbone, we evaluate several ViT architectures of different scales: ViT-Tiny, ViT-Small, and ViT-Base \cite{touvron2021training}. Table~\ref{tab:ablation_backbone} details the different ViT network architectures and their corresponding clustering results. For this ablation study, we default the mask ratio to 0.5 and maintain the training strategy described in Section~\ref{sec:training strategy}. For smaller datasets, such as RSOD, we observe that the ViT-Base runs with masked random sampling may be unstable. Nevertheless, the ViT-Small architecture consistently outperforms both ViT-Tiny and ViT-Base on the RSOD and Chaoyang datasets. Empirically, when using ViT-Small as the backbone, our proposed PICI approach is adept at learning robust representations while striking a balance between effectiveness and efficiency.

\subsubsection{Influence of PISD and CLI}
\begin{table}[!t]\scriptsize
	\caption{The influence of PISD and CLI.}
	\label{tab:ablation_pir_and_ci}
	\renewcommand
	\arraystretch{1.25}
	\centering
	\scalebox{1}{
		\begin{threeparttable}	\begin{tabular}{m{1.2cm}<{\centering}|m{1.5cm}<{\centering}m{1.5cm}<{\centering}|m{0.8cm}<{\centering}m{0.6cm}<{\centering}m{0.6cm}<{\centering}}
				\toprule
				Dataset  &PISD &CLI  &NMI &ACC &ARI\\
				\midrule
				\multirow{3}*{RSOD}
				&\ding{55} &\ding{55}  &0.489 &0.550 &0.409\\
				&\ding{51} &\ding{55}  &0.527 &0.564 &0.434\\
				&\ding{51} &\ding{51}  &\textbf{0.583} &\textbf{0.772} &\textbf{0.510}\\
				\midrule
				\multirow{3}*{Chaoyang}
				&\ding{55} &\ding{55}  &0.211 &0.490 &0.188\\
				&\ding{51} &\ding{55} &0.278 &0.533 &0.263\\
				&\ding{51} &\ding{51}  &\textbf{0.382} &\textbf{0.595} &\textbf{0.382}\\
				\bottomrule
			\end{tabular}
	\end{threeparttable}}\vskip 0.1in
\end{table}

In our PICI model, we incorporate an image reconstruction task that employs ViT as the backbone combined with masked random sampling in the PISD module, and seek to cultivate more reliable and discriminative representations by enabling mutual interaction between instance-level and cluster-level spaces in the CLI module. To assess the impact of PISD and CLI, we exclude each in turn to ascertain their individual significance. When PISD is excluded, we substitute the CNN (e.g., ResNet34) with ViT as the backbone, which only consists of ResNet and PICD and can be regarded as a vanilla CC \cite{li2021contrastive}. Evidently, the performance achieved by only employing PISD surpasses that of vanilla CC, implying that ViT offers superior representation learning capability compared to CNN. Furthermore, integrating both PISD and CLI yields more remarkable clustering results than using PISD only, which confirm the contribution of both the PISD and CLI modules.

\subsubsection{Influence of Contrastive Projectors}
\begin{table}[!t]\scriptsize
	\caption{The influence of PICD.}
	\label{tab:ablation_pic}
	\renewcommand
	\arraystretch{1.25}
	\centering
	\scalebox{1}{
		\begin{threeparttable}	\begin{tabular}{m{1.2cm}<{\centering}|m{1.5cm}<{\centering}m{1.5cm}<{\centering}|m{0.8cm}<{\centering}m{0.6cm}<{\centering}m{0.6cm}<{\centering}}
				\toprule
				Dataset  &Instance-MLP &Cluster-MLP  &NMI &ACC &ARI\\
				\midrule
				\multirow{3}*{RSOD}
				&\ding{51} &\ding{55}  &0.342 &0.544 &0.159\\
				&\ding{55} &\ding{51}  &0.293 &0.524 &0.244\\
				&\ding{51} &\ding{51}  &\textbf{0.441} &\textbf{0.546} &\textbf{0.367}\\
				\midrule
				\multirow{3}*{Chaoyang}
				&\ding{51} &\ding{55}  &0.307 &0.431 &0.201\\
				&\ding{55} &\ding{51} &0.124 &0.432 &0.108\\
				&\ding{51} &\ding{51}  &\textbf{0.351} &\textbf{0.568} &\textbf{0.331}\\
				\bottomrule
			\end{tabular}
	\end{threeparttable}}\vskip 0.1in
\end{table}

In the PICD module, we enforce contrastive learning with two contrastive projectors, namely, the instance-MLP for the instance contrastive loss and the cluster-MLP for the cluster contrastive loss. To evaluate the efficacy of these two projectors (corresponding to two contrastive losses, respectively), we perform ablation studies by excluding one projector at a time and then training the model from scratch. When the cluster-MLP is excluded, we simply apply the $K$-means method to the representations produced by the instance-MLP to obtain the final cluster assignments. As depicted in Table~\ref{tab:ablation_pic}, jointly employing both instance-MLP and cluster-MLP consistently yields superior clustering performance (w.r.t. NMI, ACC, and ARI) compared to using just one projector. Specifically, on the RSOD dataset, only using the instance-MLP leads to an NMI of 0.342, while only employing the cluster-MLP yields a lower NMI of 0.293. In terms of the NMI on the Chaoyang dataset, the performance of the cluster-MLP is much better than that of the instance-MLP. To summarize, we have two observations from the experimental results. First, the instance-MLP can usually learn more discriminative representations than the cluster-MLP. Second, the joint incorporation of these two contrastive projectors can provide more robust clustering performance than using one projector only.

\begin{figure*}[!t]
	\begin{center}
		{\subfigure[RSOD]
			{\includegraphics[width=0.45\columnwidth]{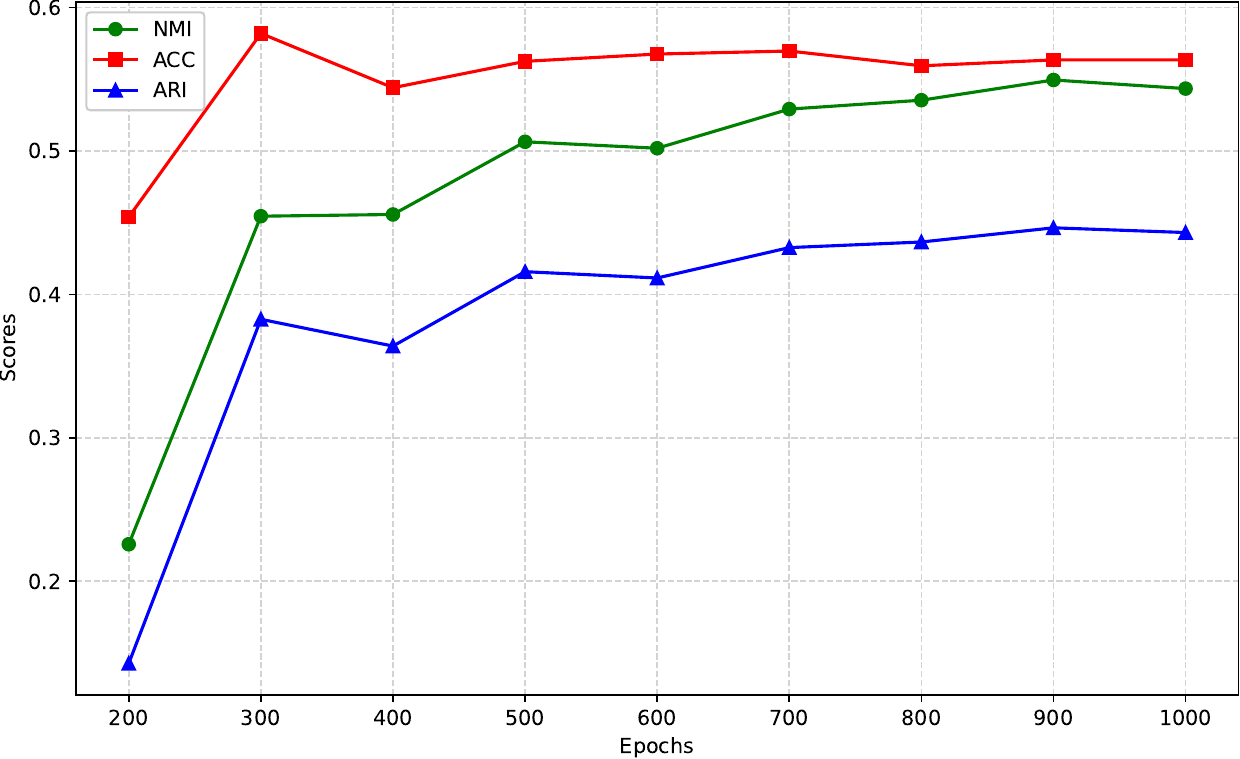}}}\hskip 0.2 in 
		{\subfigure[Chaoyang]
			{\includegraphics[width=0.45\columnwidth]{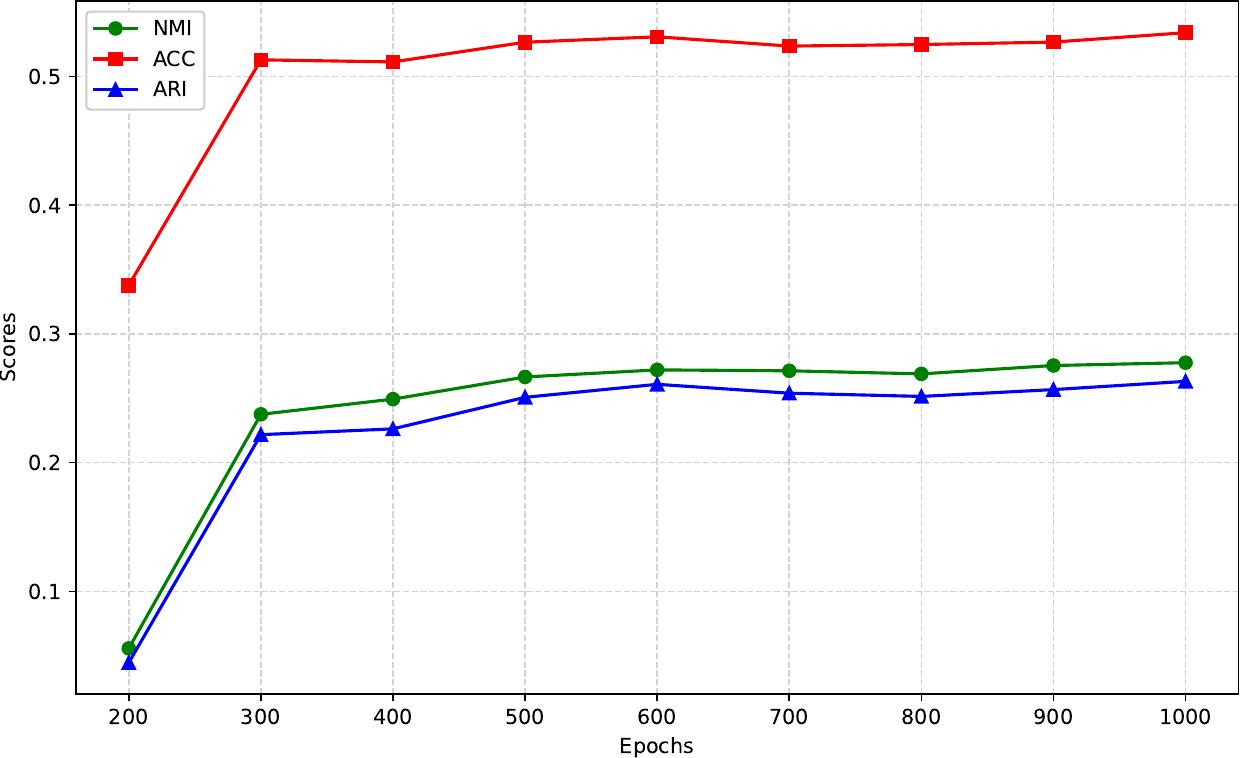}}}\vskip -0.08 in
		\caption{Illustration of the convergence of PICI (w.r.t. NMI, ACC, ARI) on the RSOD and Chaoyang dataset.}\vskip -0.08in
		\label{fig:convergences}
	\end{center}\vskip -0.15 in
\end{figure*}

\begin{figure*}[!t]\vskip 0.1in
	\centering
	\subfigure[0 epoch]{
		\includegraphics[width=0.23\textwidth]{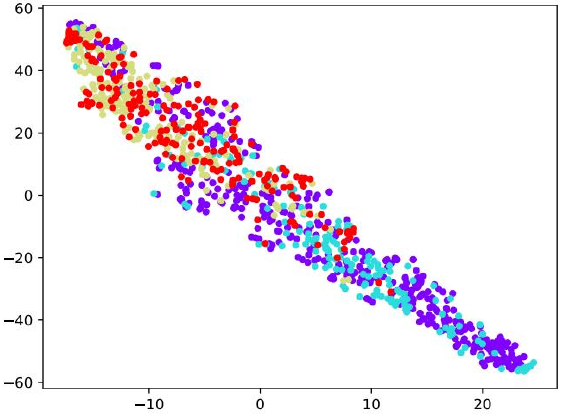}}
	\subfigure[200 epoch]{
		\includegraphics[width=0.23\textwidth]{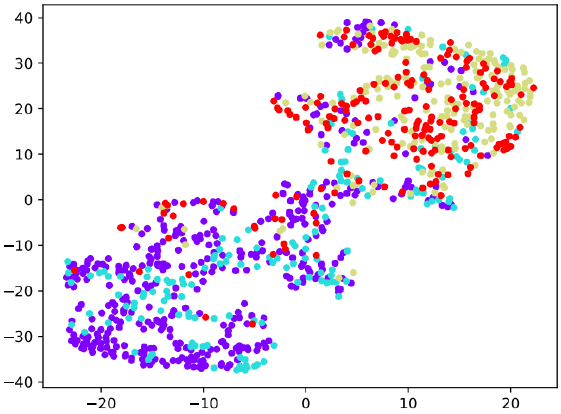}}
	\subfigure[1000 epoch]{
		\includegraphics[width=0.23\textwidth]{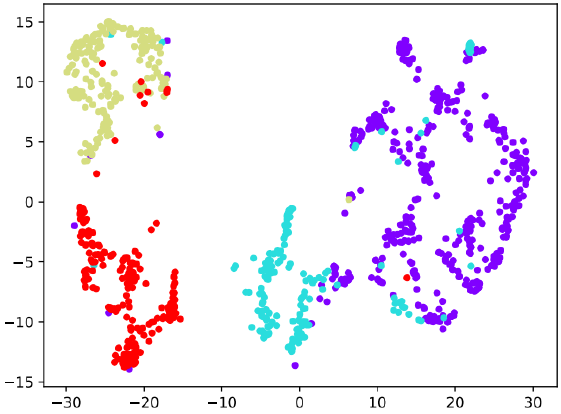}}
	\subfigure[final epoch]{
		\includegraphics[width=0.23\textwidth]{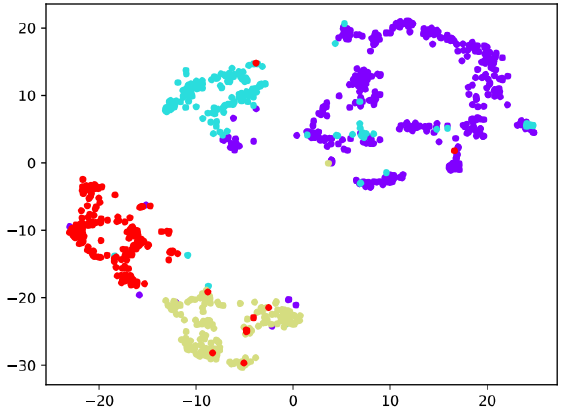}}
	\caption{The t-SNE visualization of PICI on the RSOD dataset.}
	\label{fig:t-SNE}
\end{figure*}

\subsection{Convergence Analysis}

In this section, we evaluate the convergence of the proposed PICI as the number of epochs increases. Specifically, the clustering scores (w.r.t. NMI, ACC and ARI) on the RSOD and Chaoyang datasets are recorded every 100 epochs in Fig.~\ref{fig:convergences}. The NMI, ACC and ARI scores rocket simultaneously at the first 200 epochs on the benchmark datasets. Subsequently, the scores of PICI consistently rise up as the number of epochs accumulates and finally reach a plateau.

Furthermore, we apply t-SNE \cite{van2008visualizing} on the representations learned by instance-MLP to visualize the convergence of PICI. Each instance-level feature is marked with different colors according to cluster assignments predicted by cluster-MLP. As shown in Fig.~\ref{fig:t-SNE}, the distribution of the features is chaotic but relatively uniform at the end of masked image pre-training. As the training process goes on, features distribute more distinguishable and balanced in instance-level subspace. Ultimately, the boosting algorithm enables the proposed PICI to achieve improved intra-cluster compactness and well-separated clusters.

In this section, we assess the convergence of the proposed PICI method with an increasing number of epochs. Specifically, we report the clustering scores (w.r.t. NMI, ACC, and ARI) on the RSOD and Chaoyang datasets every 100 epochs, as illustrated in Fig.~\ref{fig:convergences}. The NMI, ACC, and ARI scores experience a sharp increase during the initial 200 epochs on the benchmark datasets. Following this, the scores of PICI consistently improve with more epochs and eventually stabilize.


\section{Conclusion}
\label{sec:conclusion}

In this paper, we present a novel deep image clustering approach termed PICI, which enforces the partial information discrimination and the cross-level interaction in a joint learning framework. In particular, we leverage a Transformer encoder as the backbone, through which the masked image modeling with two paralleled augmented views is formulated. After deriving the class tokens from the masked images by the Transformer encoder, three partial information learning modules are further incorporated, including the PISD module for training the auto-encoder via masked image reconstruction, the PICD module for employing two levels of contrastive learning, and the CLI module for mutual interaction between the instance-level and cluster-level subspaces. Extensive experiments have been conducted on six real-world image datasets, which demononstrate the superior clustering performance of the proposed PICI approach over the state-of-the-art deep clustering approaches.

\bibliographystyle{elsarticle-num-names}
\bibliography{refs}

\newpage

\end{document}